\def\year{2021}\relax
\documentclass[letterpaper]{article} 
\usepackage{aaai21}  
\usepackage{times}  
\usepackage{helvet} 
\usepackage{courier}  
\usepackage[hyphens]{url}  
\usepackage{graphicx} 
\usepackage{natbib}  
\usepackage{caption} 
\usepackage{tabularx}
\usepackage{booktabs}	
\usepackage{microtype}
\usepackage{siunitx}    
\usepackage{amsmath,amsfonts,amssymb,amsthm} 
\usepackage{color, soul}    
\title{From Hesitancy Framings to Vaccine Hesitancy Profiles: A Journey of Stance, Ontological Commitments and Moral Foundations}

\author{
  Maxwell Weinzierl, \textsuperscript{\rm 1}
  Sanda Harabagiu \textsuperscript{\rm 1}\\
}
\affiliations{
    \textsuperscript{\rm 1}Human Language Technology Research Institute, The University of Texas at Dallas\\
    \{maxwell.weinzierl, sanda\}@utdallas.edu
}

\begin{document}

\maketitle

\begin{abstract}
While billions of COVID-19 vaccines have been administered,
too many people remain hesitant. Twitter, with its substantial reach and daily exposure, is an excellent
resource for examining how people frame their vaccine hesitancy and to uncover vaccine hesitancy profiles.
In this paper we expose our processing journey from identifying Vaccine Hesitancy Framings in a collection
of 9,133,471 original tweets discussing the COVID-19 vaccines, establishing their ontological commitments,
annotating the Moral Foundations they imply to the automatic recognition of the {\em stance} of the tweet
authors toward any of the {\sc CoVaxFrames} that we have identified. When we found that 805,336 Twitter users had a stance towards some {\sc CoVaxFrames} in either the 9,133,471 original tweets or their
17,346,664 retweets, we were able to derive nine different Vaccine Hesitancy Profiles of these users and to interpret these profiles based on the ontological commitments of the frames they evoked in their tweets and on value of their stance towards the evoked frames.

\end{abstract}

\section{Introduction}
Social media microblogging platforms, specifically Twitter, have become highly influential and relevant to 
shaping attitudes towards vaccination. With 206 million daily active users as of 2021, Twitter has substantial reach and daily exposure, being the most popular social network for news consumption \cite{pew1}. Since 
Twitter allows people to express their beliefs about vaccines and their hesitancy to vaccinate, their trust or mistrust in vaccines as well as their stance on vaccination mandates, it is an excellent
resource for investigating how vaccine hesitancy is framed. While vaccine hesitancy 
is mostly believed to be fueled by misinformation \cite{Renee_Sean}, in our study of
the Twitter discourse focusing on the COVID-19 vaccines, we have found that misinformation is not 
the only explanation for vaccine hesitancy. Vaccine hesitancy is also driven by the erosion of trust in vaccines, even when no misinformation is referred, or by lack of health literacy and even by the interaction between civil rights and vaccination mandates.

\begin{table}[ht]
\centering
\small
\begin{tabular}{p{0.45\textwidth}}
    \toprule \toprule
    {\bf Vaccine Hesitancy Framing 1:} {\em Governments hide vaccine safety information.} \\
    \midrule
    {\sc Stance:} {\bf Accept}\\
    {\em Tweet:} Why would the government block the Office for National Statistics from publishing side effects and deaths after taking covid vaccine? What are they hiding?
    \\ 
    \hline
    {\sc Stance:} {\bf Reject} \\
    {\em Tweet:} @USER I am not talking about protection, but prima facie what i have heard from many in government and private hospital doctors is, 2 doses are very much effective in stopping mortality or high organ damage. Vaccine can't stop covid, we are talking more about mortality reduction
    \\ \hline \hline
    {\bf Vaccine Hesitancy Framing 2:} {\em It is not known if the COVID-19 vaccines will provide protection against future variants.} 
    \\ \hline
    {\sc Stance:} {\bf Accept}\\
    {\em Tweet:} It’s not a vaccine. The COVID-19 mRNA vaccine does not provide immunity to Covid or it’s variants so you can still catch Covid and transmit it to others making you asymptomatic. You will likely need a booster shot every 6 months, so get ready to roll up that sleeve every six mon
    \\ \hline
    {\sc Stance:} {\bf Reject}\\
    {\em Tweet:} @USER @USER Reasons to get the vaccine: 1. It can protect you in case your immune system can't fight the virus. 2. It can help protect your community and vulnerable people. 3. It can help to prevent the spread of variants. 4. We don't know the long-term effects of COVID-19 yet \#GetTheShot
    \\ \hline
    \bottomrule
\end{tabular}
\caption{Examples of Vaccine Hesitancy Framings and tweets evoking them, while the tweet authors accept or reject the framing.}
\label{tb:VHFs}
\end{table}

\begin{figure*}[t]
    \centering
    \includegraphics[width=0.98\linewidth]{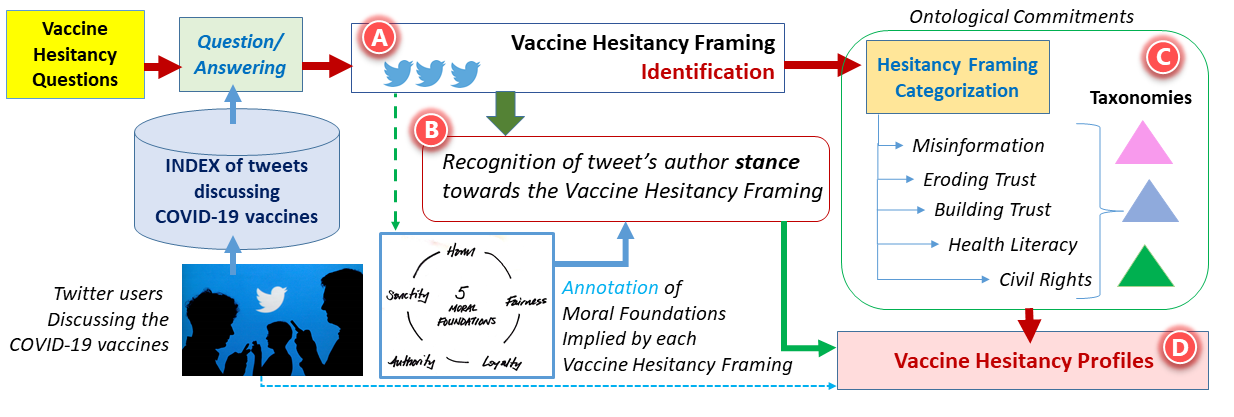}
    \caption{Recognizing Vaccine Hesitancy Profiles (VHPs) by taking into account (1) the identification of Vaccine Hesitancy Framings (VHFs); (2) the {\em stance} of tweet authors towards the framings; and (3) the ontological commitments of the identified framings.}
    \label{fig:system}
\end{figure*}

Social science stipulates, according to \citet{Druckman, Entman},  that discourse almost inescapably involves framing – a strategy of highlighting certain issues to promote a certain interpretation or attitude. 
{\bf Vaccine Hesitancy Framings} (VHFs), highlight issues regarding  
confidence in the safety of vaccines by using {\em specific} misinformation, as exemplified by the first VHF listed in Table~\ref{tb:VHFs}, or erode the trust in vaccines by demotivating people from vaccination, as exemplified in the second VHF listed in Table~\ref{tb:VHFs}. While VHFs are not directly expressed in tweets discussing vaccines, they can be inferred from the discourse spanning these tweets. Furthermore, tweet authors do not only evoke these VHFs, but they also express their {\em stance} towards them. Table~\ref{tb:VHFs} illustrates a tweet whose author agrees with the first VHF shown in the Table, thus {\em adopting} the framing and another tweet whose author disagrees with the same framing, thus {\em rejecting} it. Tweets with different stance for the second VHF are listed as well in the Table.

Vaccine hesitancy is a continuum between those that accept vaccines with no doubts, to those that absolutely refuse vaccines, with vaccine hesitant individuals in heterogeneous groups between these two extremes, according to \cite{SAGE}. 
To uncover the {\bf Vaccine Hesitancy Profiles} (VHPs), which was the main objective of our study, we focused on the Twitter discourse regarding the COVID-19 vaccines. 
Intuitively, all tweet authors sharing the same VHP are expected to also share some commonalities through the way they frame their hesitancy towards COVID-19 vaccines. 
Therefore, as illustrated in Figure~\ref{fig:system}, our methodology considered (A) the identification of VHFs from an index of 5,865,046 tweets discussing COVID-19 vaccines and 
(B) the recognition of the stance towards the VHFs evoked by 805,336 users. 
Furthermore, we relied on previous work in social psychology considered the Moral Foundations Theory (MFT) \cite{Haidt_Graham, Haidt_Joseph} as a theoretical framework for analyzing moral framing, using the same five key values of human morality, emerging from evolutionary, social, and cultural origins. 
Each VHF was annotated with the Moral Foundation (MF) they imply. For example, for the first VHF illustrated in Table~\ref{tb:VHFs}, the annotated MFs are: \emph{Harm}, \emph{Betrayal} and \emph{Authority} while for the second VHF illustrated in the same table, the only annotated MFs is \emph{Betrayal}. The annotated MFs informed the recognition of the tweet author stance towards the VHFs 
that are evoked in their tweets. These annotations have contributed to the automatic recognition of stance.

Figure~\ref{fig:system} shows that our journey, from identifying VHFs (revealed as answers to questions about vaccine hesitancy), to finally recognizing the VHPs, 
involved also (C) the derivation of the ontological commitments of the VHFs, by categorizing them into
misinformation, framings eroding trust in vaccines, framing building trust in vaccines, framings showcasing the health literacy of the tweet author or framings in which civil rights are brought up. 
This categorization enabled us to identify the common themes and concerns of the VHFs in each category, and to organize them into taxonomies, completing the ontological organization of the VHFs. 
These taxonomies, together with the stance information, allowed us to create a representation of each Twitter user that 
we recognized to have a stance towards the VHFs, and (D) to reveal the VHPs. 
The ontological information along with the stance information enables us to interpret the VHPs. 
For example, the tweet authors mostly evoking the first VHF illustrated in Table~\ref{tb:VHFs} belong to the profile of {\sc Undecided}, whereas the authors accepting the second VHF from the same table are {\sc Demotivated} in their hesitancy, whereas those rejecting it belong to the profile of those that are {\sc Motivated} to vaccinate.

\section{Identification of Vaccine Hesitancy Framings through Question/Answering} 
\label{sec:vhfs}

\begin{figure*}
    \centering
    \includegraphics[width=0.91\linewidth]{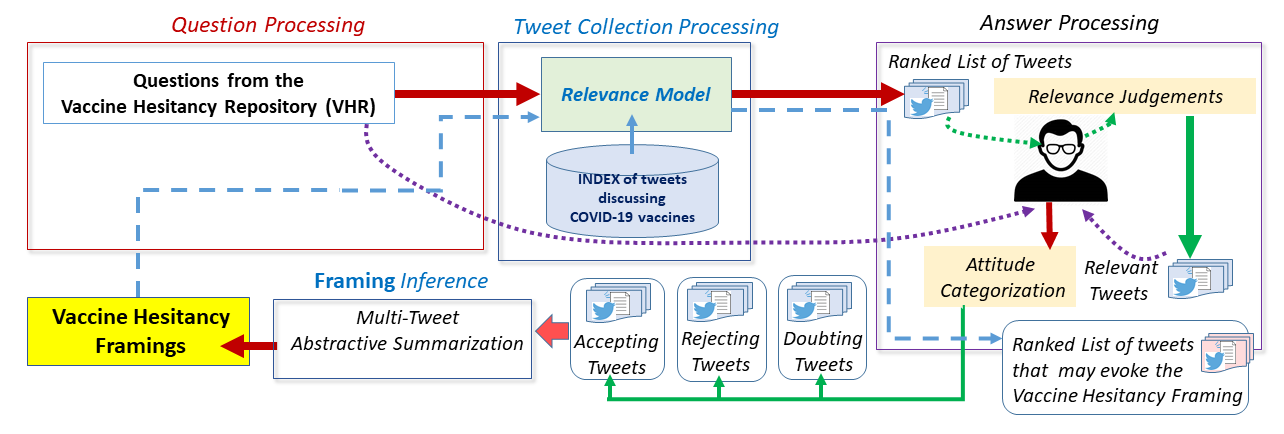}
    \caption{A Question Answering Framework for (a) identifying Vaccine Hesitancy Framings as well as (b) tweets that potentially evoke a Vaccine Hesitancy Framing.}
    \label{fig:QA}
\end{figure*}

Current NLP methods \cite{TenYears,TaoCui} used for recognizing vaccine hesitancy assume that a neutral sentiment detected in a tweet is equivalent to hesitancy,
while a positive sentiment is interpreted as acceptance of vaccination and a negative sentiment as refusal. However, as defined in \cite{SAGE},
vaccine hesitancy refers to the delay in acceptance or refusal of vaccines despite availability of vaccination services. Moreover, vaccine hesitancy is informed by factors such as complacency, convenience, and confidence, which are framed in complex ways in language \cite{3C}.
For example, when misinformation is used in framing vaccine confidence, it typically results in vaccine hesitancy. Similarly, when civil rights are highlighted in a particular framing, it promotes vaccine refusal, while when trust in vaccines is increased, it leads to vaccine acceptance, and eventual uptake.
Therefore, developing novel NLP techniques capable of discovering framings of vaccine hesitancy in social media discourse is essential, especially for uncovering the various vaccine hesitancy profiles of users. 
 
Recent work in NLP concerning automatic recognition of framings targeted the study of political bias and polarization in social and news media \cite{field-etal-2018-framing,roy-goldwasser-2020-weakly}, mainly addressing the recognition of 15 cross-cutting dimensions of political framings e.g., economic dimensions, fairness and equality or policy prescription and evaluation. Although recent Twitter content analysis \cite{Lerman20} revealed that there is significance correlation between polarized attitudes towards vaccines and political dimensions, to our knowledge, no NLP methods have yet been developed to identify vaccine hesitancy framings, although vaccine hesitancy is often discussed in social/news media.

In a novel approach that uses Question Answering, illustrated in Figure~\ref{fig:QA}, we have found that VHFs focusing on the COVID-19 vaccines can be successfully identified as answers to questions from the Vaccine Confidence Repository (VCR) \cite{Rossen}, which is a set of 18 questions targeting hesitancy, informed by the anti-vaccine content analysis reported in \cite{Kata}. 
The same questions were used in the study reported in \cite{Rossen} to discern hesitancy profiles from the answers returned on survey links available from Facebook pages and parenting forums. 
The questions from VCR targeted the Human Papillomavirus (HPV) vaccine. 
We adapted the questions by asking about the COVID-19 vaccine. 
These questions covered five different themes stipulating that: (T1) vaccines are unsafe and unnatural; (T2) vaccination is ineffective; (T3) redundant vaccinations; (T4) people should be free to decide if they 
want to vaccinate; and (T5) vaccination is a conspiracy. 
Instead of soliciting answers from Twitter users, we decided to (a) automatically find tweets that answer the same questions and (b) infer the framings evoked by the answers. 

As shown in Figure~\ref{fig:QA} each question was processed, transforming it into a query that can be handled by a relevance model implementing the 
BM25 vector relevance model \cite{bm25}. 
In addition, as in \cite{ser4eqnova}, we considered the BERT-RERANK \cite{bert-rerank} 
scoring function to re-rank the tweets provided by the BM25 relevance model. 
But, finding the answers to the questions required an index of tweets discussing the COVID-19 vaccines. 

In order to obtain a collection of tweets discussing the COVID-19 vaccine, we started by obtaining
approval from the Institutional Review Board at  
the University of Texas at Dallas.
IRB-21-515 stipulated that our research met the criteria for exemption \#8(iii) of the Chapter 45 of Federal Regulations Part 46.101.(b). 
Afterwards, tweets discussing the COVID-19 vaccines were obtained by using the query {\em ``(covid OR coronavirus) vaccine lang:en”}. 
A collection of 9,133,471 original tweets and 17,346,664 retweets was obtained from the Twitter streaming API. 
These tweets were authored between December 18th, 2019, and July 21st, 2021.
To detect duplications in the original tweets, we performed Locality Sensitive Hashing (LSH) \cite{lsh} with term trigrams, 100 permutations, and a Jaccard threshold of 50\%, obtaining 5,865,046 unique original tweets discussing COVID-19 vaccines. 
We first build an index for 5,865,046 unique original tweets using Lucene\cite{lucene}, informing the relevance model.



\begin{figure*}
    \centering
    \includegraphics[width=0.91\linewidth]{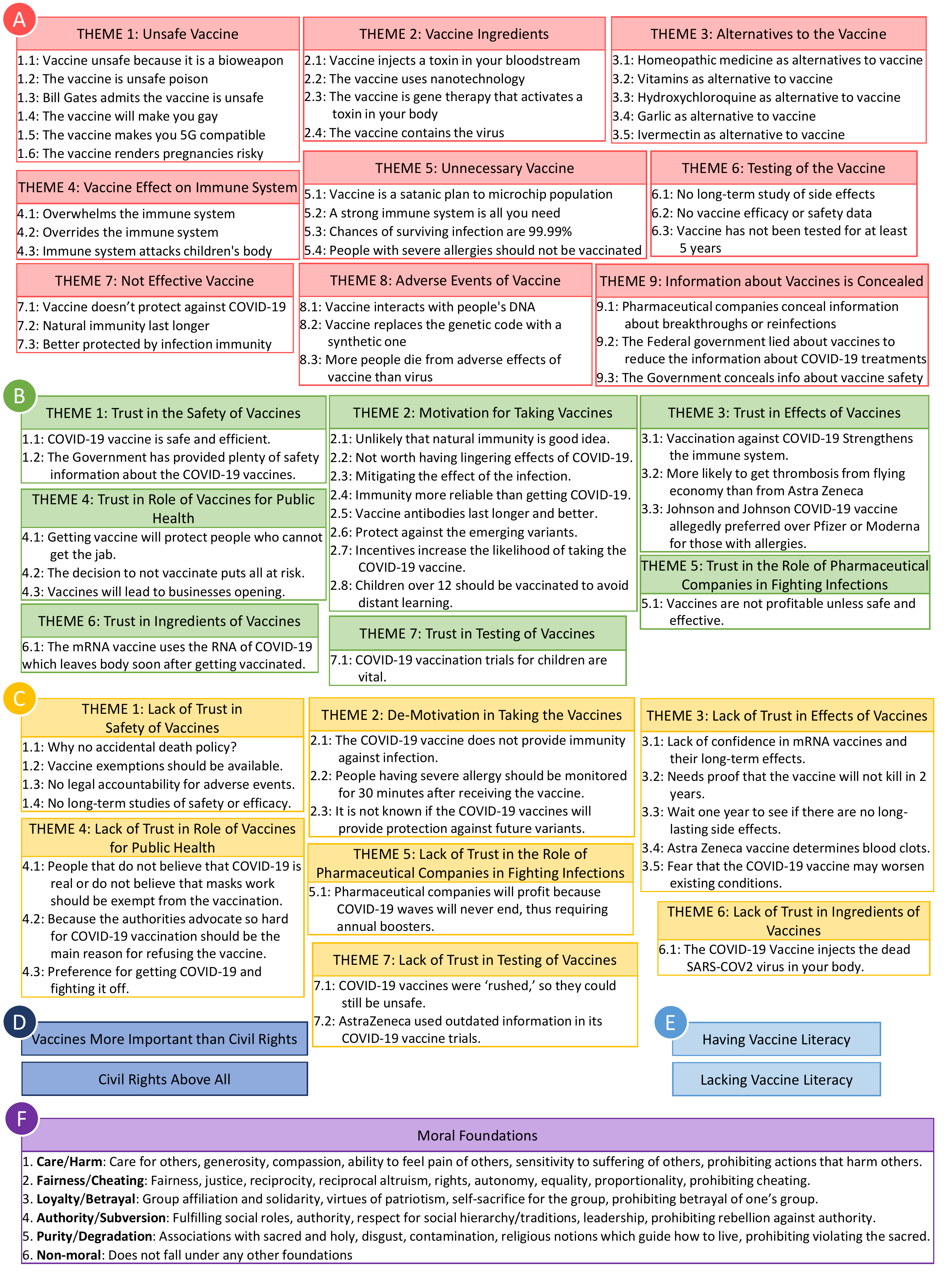}
    \caption{Taxonomy of Themes and Concerns of (A) Misinformation, (B) Building Trust, (C) Eroding Trust, (D) Civil Rights, (E) Vaccine Literacy, and (F) Moral Foundations defined by the Moral Foundation Theory, annotated when implied in Vaccine Hesitancy Framings from {\sc CoVaxFrames}.}
    \label{fig:hierarchy}
\end{figure*}

Relevance judgements were produced on the 300 best ranked tweets from the first index, language experts categorizing afterwards the attitude against the predication of the question of each relevant tweet. 
Less than 60\% of the tweets were judged relevant by two language experts from the University of Texas at Dallas.
As shown in Figure~\ref{fig:QA}, relevant tweets were categorized as accepting the predication, rejecting it and doubting it, with a Cohen Kappa score of 0.81, which indicates strong agreement
between annotators (0.8-0.9) \cite{kappa}.
From tweets sharing the same attitude towards the question predication, the Pyramid method \cite{pyramid} was used to infer a query-focused multi-tweet summary, which 
was considered a VHF. In this way, we identified a set of 113 VHFs targeting the COVID-19 vaccine, which we assembled in {\sc CoVaxFrames}.  Then, each VHF
from {\sc CoVaxFrames} was reused as a question and processed by the relevance model against a new
index containing all the 9,133,471 original tweets and 17,346,664 retweets.
From the best-ranked 400,000 tweets retrieved for each VHF, only the tweets that had a relevance score above a threshold $T_r=2.0$, selected from initial experiments, were considered to potentially evoke the VHF, which resulted in 19,233,144 tweets potentially evoking a VHF from {\sc CoVaxFrames}.



\section{Ontological Commitments of Vaccine Hesitancy Framings}
\label{sec:tax}

The VHFs were inspected first to distinguish which ones contain misinformation and which do not. 
Of the 113 VHFs from {\sc CoVaxFrames}, 38 VHFs were categorized as Misinformation framings. The remaining VHFs were categorized in the following way: 52 VHFs addressed issues of trust in vaccines, but without using misinformation; 32 VHFs addressed issues of CIVIL Rights. In addition, we inspected all VHFs to decide if 
they exhibit health literacy or lack of it: 28 VHFs were categorized as Literacy framings. For the VHFs
addressing issues of trust in vaccines, we found that 27 VHFs are Eroding Trust in vaccines while  25 VHFs are Building Trust framings. 
This categorization allowed us to organize a Misinformation Taxonomy, which encodes the common themes and concerns of the Misinformation framings. Similarly, we have organized a taxonomy of Building Trust and a
taxonomy of Eroding Trust. These taxonomies are illustrated in Figure~\ref{fig:hierarchy}. 
For the VHFs addressing Civil Rights issues, we have created only two themes, as we have done for the
VHFs exhibiting Vaccine Literacy or lack thereof, in their respective ontologies.

In addition, all VHFs from {\sc CoVaxFrames} were annotated with as many Moral Foundations as they implied. Figure~\ref{fig:hierarchy} (F) lists the definitions of each MF from the Moral Foundation Theory \cite{Haidt_Joseph} that we have used. 
A computational linguist and an expert in public health have independently assigned MFs to all the VHFs, and the inter-judge agreement was a Cohen's Kappa score of 0.85, where disagreements were resolved between annotators.
The most common Moral Foundations were \emph{Harm} and \emph{Subversion}, occurring in 45 VHFs and 44 VHFs respectively out of the 113 VHFs, while the least common Moral Foundations were \emph{Cheating} and \emph{Loyalty}, occurring in 8 VHFs and 9 VHFs respectively. 
These ontological commitments that were organized in the {\sc CoVaxFrames} contributed to the discovery of hesitancy profiles, along with the information about the stance towards any of the VHFs evoked in tweets.

\section{Stance Recognition}
\label{sec:model}

\begin{figure}[ht]
    \centering
    \includegraphics[width=0.98\linewidth]{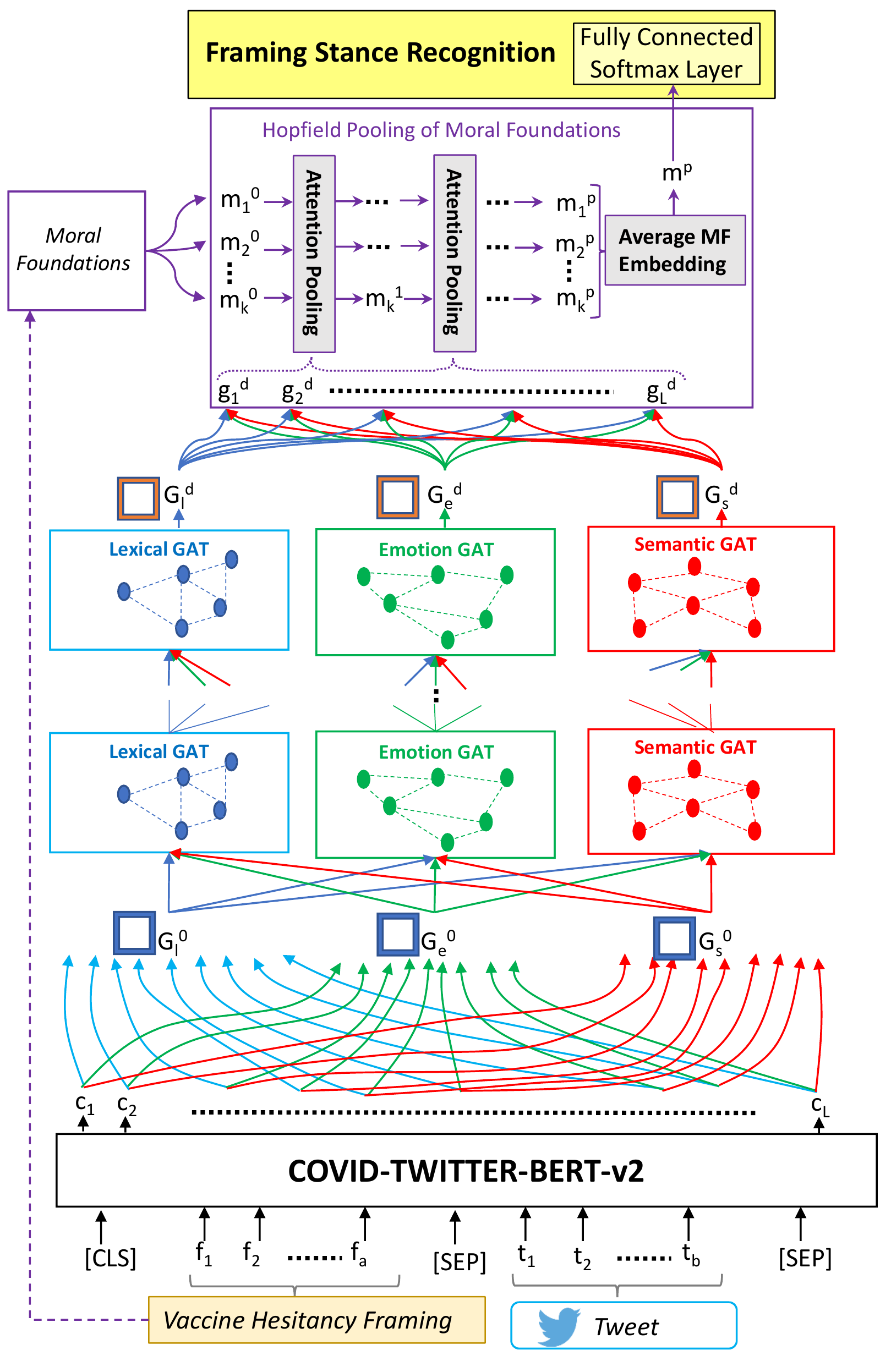}
    \caption{Neural architecture of {\sc StanceId-Morality}.}

    \label{fig:architecture}
\end{figure}

The recognition of the stance of a tweet author towards any of the  {\sc CoVaxFrames} is made possible 
by the {\sc StanceId-Morality} system, illustrated in Figure~\ref{fig:architecture}. 
Given any VHF $f$ from {\sc CoVaxFrames} and any tweet $t$ that may evoke a VHF, produced by the Q/A framework illustrated in Figure~\ref{fig:QA}, we hypothesize that if an {\em Accept} or {\em Reject} stance towards VHF $f$ is recognized automatically, then the tweet $t$ is recognized as evoking VHF $f$, 
otherwise tweet $t$  does not evoke the VHF $f$. We also believed that the stance of the author
of tweet $t$ is revealed not only by the interactions between lexical, semantic and emotion information
expressed in the tweet $t$, but also by their interactions with the Moral Foundations (MFs) implied by the VHF $f$. Therefore, we 
designed for the {\sc StanceId-Morality} system
a novel neural architecture that combines the advantages of the contextual embeddings learned by COVID-Twitter-BERT-v2 \cite{covid-twitter-bert} with the Graph Attention Networks (GATs) \cite{gat} where lexical, emotion, and semantic information can be processed \cite{covid-misinfo-stance} and a special case of Continuous Hopfield Networks \cite{HopfieldNets}, namely Hopfield Pooling, where the MFs can be processed as well.

The processing of a  VHF $f$ from {\sc CoVaxFrames} and the tweet $t$ that potentially evokes $f$ starts in the {\sc StanceID-Morality} system with  joint 
word-piece tokenization 
\citep{bert}, producing the sequence of word-piece tokens $[CLS], f_1, f_2, ..., f_a, [SEP], t_1, t_2, ..., t_b, [SEP]$. This sequence of tokens is provided to COVID-Twitter-BERT-v2 for  generating the corresponding contextualized embeddings.

COVID-Twitter-BERT-v2 is a language model which was pre-trained on 97M COVID-19 tweets, providing domain-specific language modeling for tasks concerning COVID-19.
All contextualized embeddings $c_1, c_2, ..., c_L$ are organized in a matrix $G \in \mathbb{R}^{L \times 1024}$, which is provided as input to either the stacked lexical GATs, the stacked semantic GATs or the stacked emotion GATs. Each of these GATs operate on graphs in which the nodes are words from $f$ or $t$.  
The lexical graph relies on dependency parse edges between words, the emotion edges connect words which share emotion tags from SenticNet 5 \citep{senticnet-5}, and the semantic edges are between pairs of words identified as semantically similar in SenticNet 5.
Each GAT operates on one of these graphs, refining the representations of each word from each graph through self-attention between adjacent connections. There are $d$ layers of GATs in {\sc StanceId-Morality}.
A GAT at layer $n\in \{1, ..., d\}$ computes a hidden representation for every contextual  embedding $g_i^{n-1}\in G^{n-1}$ as $h_i^n = W^n g_i^{n-1}$, where $W^n$ is a learned weight matrix. This hidden representation is required for computing the self-attention weights of each GAT:
\begin{equation}
    \alpha^n_{i,j} = \frac{exp(LeakyReLU((a^n)^T[h_i^n, h_j^n]))}{\sum_{k \in adj(i)}{exp(LeakyReLU((a^n)^T[h_i^n, h_k^n]))}}
\end{equation}
where $a^n$ is a learned weight vector of size $2F$, $[..,..]$; represents concatenation; $LeakyReLU(x)=max(0.2x, x)$; and $adj(...)$ produces the list of adjacent nodes for a given node from the Lexical, Emotion, or Semantic graphs. 
The attention weights $\alpha^n_{i,j}$ determine the output of each GAT at layer $n$:
\begin{equation}
    g_i^n = \sigma({\sum_{j \in adj(i)}{\alpha^n_{i,j}h_j^n}}) 
\end{equation}
where $\sigma$ is an exponential linear unit (ELU) nonlinearity \citep{elu}.

Each of the $d$ layers of GATs have a hidden size $F$, producing a graph representation $G_l^n$, $G_e^n$, and $G_s^n$ respectively of size $L \times F$. 
The GAT hidden size $F$ and the number of layers $d$ are selected from experiments on the development collection, outlined in Section~\ref{sec:results}.
These Lexical, Emotion, and Semantic Graph representations are concatenated together to form $G^n=[G_l^n, G_e^n, G_s^n]$, with $G^n \in \mathbb{R}^{L\times 3F}$, which is provided as input to all three Lexical, Emotion, and Semantic GATs for the next layer, producing $G^{n+1}$. 
This allows each Lexical, Emotion, and Semantic GAT to consider previous Lexical, Emotion, and Semantic Graph representations jointly, learning graph node embeddings which consider interactions between different graphs. 
The output of the final GAT layers  $G^d=[G_l^d, G_e^d, G_s^d]$ is 
provided to the Hopfield Pooling of Moral Foundation (HP-MF) module. 

Each of the $10$ MFs $m_i$ are each assigned a unique Moral Foundation Embedding (MFE) $m_i^0 \in \mathbb{R}^{3F}$, initialized randomly and learned throughout the training process of the {\sc StanceId-Morality} system. 
The MFEs of the $k$ MFs annotated for the VHF $f$, $m_1^0, m_2^0, ..., m_k^0$, are used as initial query embeddings for performing independent Hopfield Pooling \cite{covid-hopfield-pool-event} on $G^d$. 
Hopfield pooling performs attention pooling $p$ times, where each iteration $j$ refines the query MFE $m_i^{j}$ from performing attention pooling on the outputs of the final GAT layers $G^d$ utilizing the previous MFE $m_i^{j-1}$ as the query.
For each of the $k$ MFs judged within the VHF $f$, we perform attention pooling at each step $j$, from $1$ to $p$, independently for each MF $m_i$.
Attention weights are computed using $m_i^{j-1}$ as the query against the final lexical, emotion, and semantic word embeddings $G^d=\{g_1^d, g_2^d, ..., g_L^d\}$:
\begin{equation}
    \beta^{i,j}_{x} = \frac{exp(g_x^d \cdot m_i^{j-1})}{\sum_{y=1}^{L}{exp(g_y^d \cdot m_i^{j-1})}}
\end{equation}
Where $\cdot$ represents the dot product.
These attention weights $\beta^{i,j}_{x}$, which range from $0$ to $1$, represent how closely the MFE $m_i^{j-1}$ aligns with each of the concatenated lexical, emotion, and semantic word embeddings $g_x^d \in G^d$.
The updated MFE $m_i^j$ is then computed as a weighted sum, using $\beta^{i,j}_{x}$ as the weights, over the lexical, emotion, and semantic word embeddings $g_x^d \in G^d$:
\begin{equation}
    m_i^{j} = {\sum_{x=1}^{L}{\beta^{i,j}_{x}g_x^d}}
\end{equation}
Hopfield pooling is therefore performed for each of the $k$ initial MFEs $m_1^0, m_2^0, ..., m_k^0$ found in VHF $f$ by attention pooling $p$ times over $G^d$ to iteratively construct $m_1^p, m_2^p, ..., m_k^p$. 
These final $k$ MFEs are summarized into a single fixed-length representation by taking the average MFE as $z=\frac{1}{k}\sum_{i=1}^{k}{m_i^p}$:

This embedding $z$ is provided to the stance recognition layer, which employs a fully connected layer with a softmax activation function to produce final probabilities $P(Accept \mid f, t)$, $P(Reject \mid f, t)$, and $P(No\_Stance \vee \neg Evoke \mid f, t)$, where we merge the probabilities $P(No\_Stance \mid f, t)$ and $P(\neg Evoke \mid f, t)$ into a single probability output, as a tweet with \emph{No Stance} towards VHF $f$ and a tweet which does not evoke VHF $f$ are both ignored when we perform vaccine hesitancy profiling.
The {\sc StanceId-Morality} system is trained end-to-end on the cross-entropy loss function:
\begin{equation}
    \mathcal{L} = -\sum_{(s, f, t) \in D}{log{P(s \mid f, t; \theta)}}
\end{equation}
where $s\in \{Accept, Reject, No\_Stance \vee \neg Evoke\}$, $D$ is a set of all training examples of labeled [tweet, VHF] pairs, and $\theta$ is a set of all trainable parameters from {\sc StanceId-Morality}. These parameters are optimized with ADAM \citep{adam}, a variant of gradient descent, to minimize $\mathcal{L}$.


\begin{table}[ht]
    \centering
    \small
    \begin{tabular}{l|rrrr|r}
        \toprule
        Split  & \emph{Evoke} & \emph{Accept} & \emph{Reject} & \emph{No\_Stance} & Total \\
        \midrule
        train &	8,390 &	5,241 &	1,668 &	1,481  & 10,250 \\
        dev & 941 &	567 &	211 &	163 &  1,115 \\
        test & 2,285 &	1,461 &	448 &	376 &	2,815 \\
        \hline
        Total &	11,616 &	7,269 &	2,327 &	2,020  & 14,180 \\
        \bottomrule
    \end{tabular}
    \caption{Distribution of stance values for VHFs in the Training, Development, and Test splits of {\sc CoVaxFrames}.}
    \label{tb:annotations}
\end{table}

Because the {\sc StanceId-Morality} system implements a supervised method for stance recognition, we 
relied on a training dataset, a development dataset as well as a testing dataset that allowed us to
perform experiments and collect results. Generating these datasets was made possible by the annotations performed on the tweets deemed relevant for each of the VHR questions used in the QA framework presented in
Section~\ref{sec:vhfs}. 
Researchers from 
the University of Texas at Dallas and public health experts from The University of California, Irvine
judged (a) whether a tweet evokes any of the VHFs from {\sc CoVaxFrames}; and (b) if so, they annotated the \emph{stance} of the tweet's author towards the VHF. 
14,180 tweets were judged, with 11,616 tweets evoking one or more VHFs from {\sc CoVaxFrames}. 
They were organized in [tweet, VHF] {\em pairs}, annotated with a stance value that could be {\em Accept}, {\em Reject} or {\em No Stance}.
Statistics for the number of tweets evoking a VHF, as well as of the stance their authors have towards the VHF, are provided in Table~\ref{tb:annotations}.
To evaluate the quality of judgements, 
we randomly selected a subset of 1,000 tweets (along with the VHF against which they have been judged a \emph{stance} value), which have been judged by at least two different language experts.
Inter-judge agreement was computed using Cohen's Kappa score, yielding a score of 0.67 for the \emph{stance} of tweets for COVID-19 VHFs, which indicates moderate agreement between annotators (0.60-0.79).

When we performed stance recognition with the {\sc StanceId-Morality} system on all the tweets that may evoke a VHF from {\sc CoVaxFrames}, produced by the QA system responding to any {\sc CoVaxFrames} (discussed in Section~\ref{sec:vhfs}),
we identified a total of 1,741,269 tweets from 805,336 users which held an \emph{Accept} or \emph{Reject} stance towards one or more VHF from {\sc CoVaxFrames}.

\section{Deriving Vaccine Hesitancy Profiles}
\label{sec:profile}

Revealing the VHPs from the 805,336 users having a stance towards any of the {\sc CoVaxFrames} requires first to produce a representation of each of these users that encodes knowledge about the way the users frame their vaccine hesitancy as well as the stance they have regarding it. To encode the knowledge
regarding vaccine hesitancy, we relied on the ontological commitments we have produced for {\sc CoVaxFrames}. More specifically, we decided to use the themes encoded in the taxonomies illustrated in Figure~\ref{fig:hierarchy} and considered the stance each user had towards VHFs within each theme. As shown in Figure~\ref{fig:user_vec}, 
we produced for each user a vector $u \in \mathbb{R}^H$, where $H= 27$ is the total number of themes across all taxonomies illustrated in Figure~\ref{fig:hierarchy} (i.e. 9 themes  from the Misinformation taxonomy;  7 themes from the taxonomy of Building Trust;  7 themes from the taxonomy of Eroding Trust;, and 2 themes each for Civil Rights  and Vaccine Literacy). The values of the vector
$u$ were computed by:

\begin{figure}[ht]
    \centering
    \includegraphics[width=0.95\linewidth]{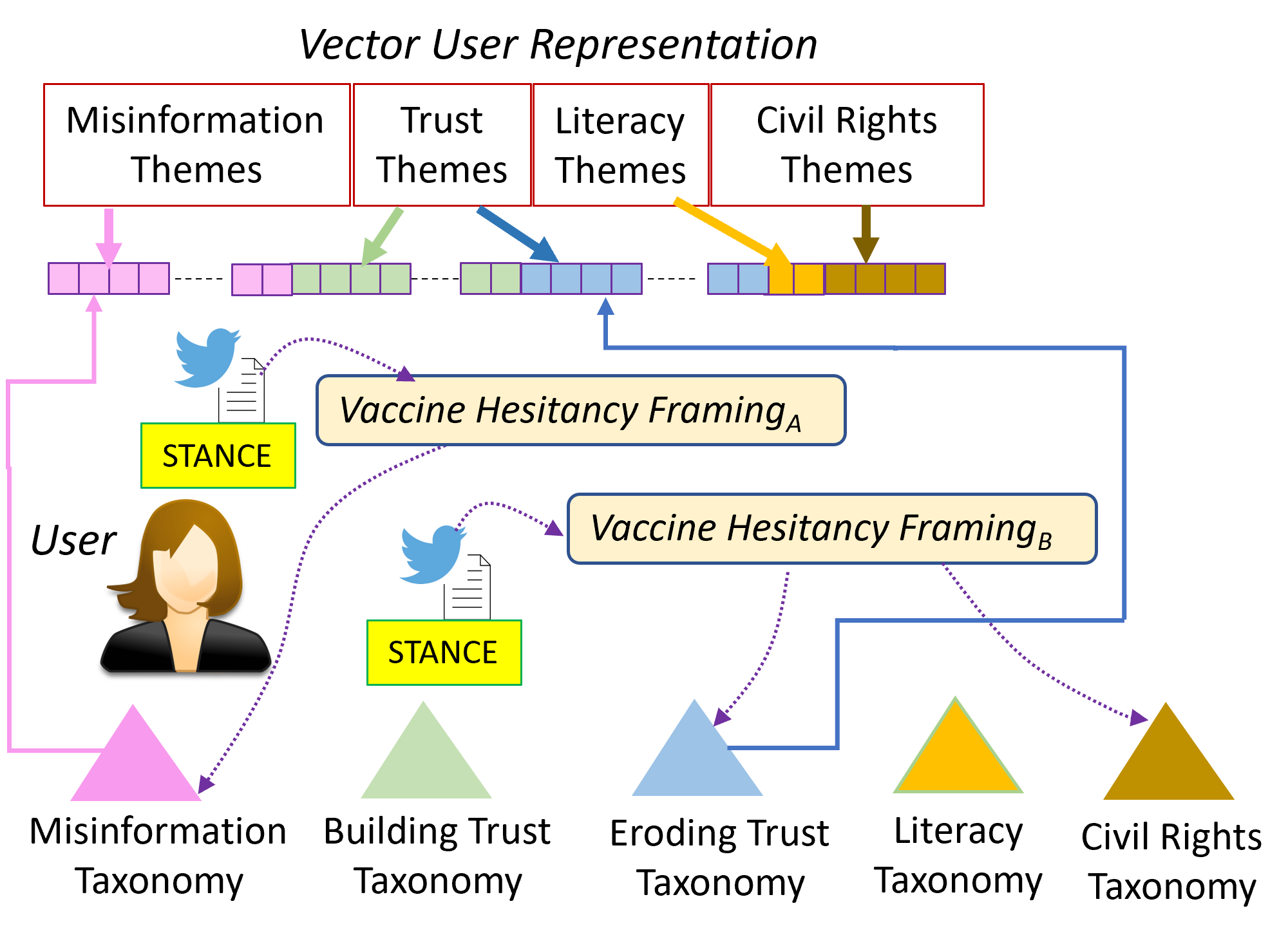}
    \caption{Representation of the Twitter users discussing vaccine hesitancy and having a stance towards hesitancy framings through Vector User Representations.}
    \label{fig:user_vec}
\end{figure}

\begin{equation} \label{eq:user_vec}
    u_{h} = \sum\limits_{(t, f, s) \in U(u, h)}\frac{SV(s)}{|U(u, h)|}
\end{equation}
where $U(u, h)$ is the set of tweets $t$ authored by $u$, evoking a VHFs $f$ belonging to the theme $h$; while $s$ represents the stance value of $u$ towards $f$, with $s \in$ \{{\em Accept, Reject}\}.  $SV(s)=1$ if the user $u$ accepted $f$ and
$SV(s)=-1$ if $u$ rejected $f$.

In this way, we obtained 805,336 sparse vectors representing the users, that enabled us to cast
the recognition of VHPs as a clustering task, which could reveal the groups of users that manifest
vaccine hesitancy with similar stance towards VHFs that share the same knowledge, encoded at the level of the hesitancy theme. 
For this purpose, we performed sparse k-means \cite{kmeans} clustering on these user vector representations, varying the number of clusters $k$ from $2$ to $12$. The final number of profiles, $k$
was selected following the Elbow method \cite{elbow}, i.e. by  (a) computing the L2-distance between the 
each VHP's centroid vector and all user vectors in the profile; and (b) computing the average distance of these distances for each VHP. We found that the number of VHPs for $k= 9$ as satisfying the Elbow method, as it obtained the minimum average distance from each user vector to the centroids of the VHPs, with an
L2-distance of $1.05$.

\begin{figure*}[th]
    \centering
    \includegraphics[width=0.99\linewidth]{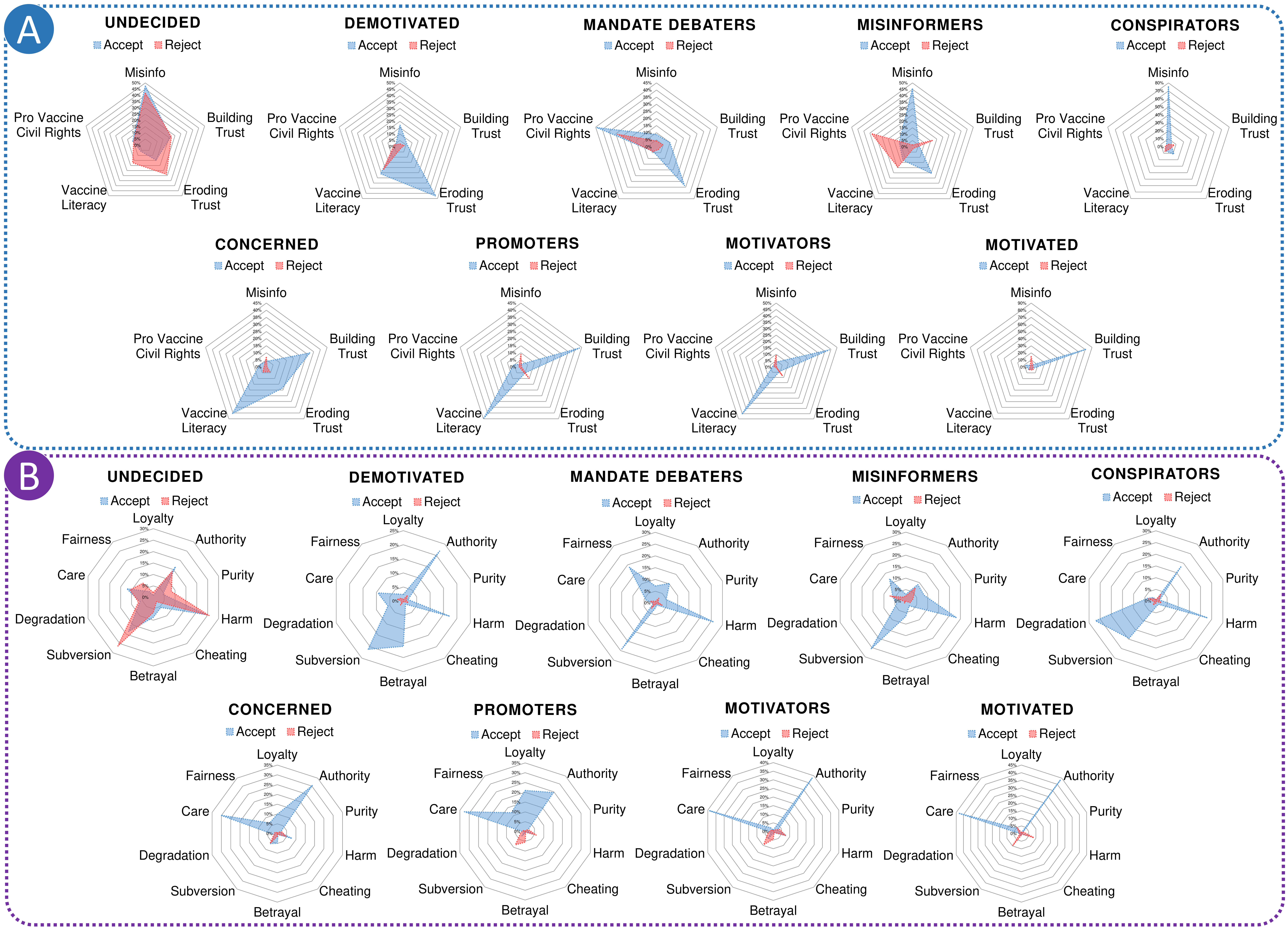}
    \caption{COVID-19 Vaccine Hesitancy Profiles (VHPs) informed by {\em stance} and (A) ontological commitments of Vaccine Hesitancy Framings (VHFs) or (B) Moral Foundations (MFs) implied by Vaccine Hesitancy Framings (VHFs).}
    \label{fig:profile_stance}
\end{figure*}

{\bf Interpretation of the Vaccine Hesitancy Profiles}\\
The $9$ VHPs were manually inspected by exploring the tweets of the $50$ users closest to the VHP centroid vectors, and each VHP was assigned a name based on the interpretation. 
Figure~\ref{fig:profile_stance} {supports our interpretations by illustrating how the predominant stances of profile users are interacting with (A) the ontology commitments; and (B) the the MFs.}

The {\underline{\sc Undecided}} VHP includes $177,836$ users (22\%) who are on the fence about the COVID-19 vaccines. 
These users are characterized by a 50/50 split in acceptance and rejection of the misinformation that the VHF that the COVID-19 vaccine is an unsafe poison, while also having a 60/40 split in trust in the government to provide accurate COVID-19 vaccine safety information.
They tend to pick-and-choose which VHFs they \emph{Accept} and \emph{Reject}, leading to theme-level inconsistencies in their beliefs. 
Users from this VHP tend to both adopt and reject VHFs with MFs of \emph{Subversion}, \emph{Harm}, \emph{Authority}, and \emph{Care}. 
They are the primary target of those that propagate COVID-19 vaccine misinformation, such as the {\sc Misinformers}, and adopt misinformation nearly as often as they reject it.

The {\underline{\sc Demotivated}} VHP includes $89,827$ users (11\%) who have largely lost their motivation to vaccinate against COVID-19. 
These users overwhelmingly accept demotivating VHFs, e.g. that the COVID-19 vaccine does not provide immunity, that you can still get infected even after getting vaccinated, and that breakthrough cases after getting fully vaccinated are common. The users in this profile
accept some COVID-19 vaccine misinformation about the vaccine ingredients, but primarily they are complacent, considering that the perceived risks of the COVID-19 vaccine do not justify the uptake. 
The predominant MFs of the VHFs they evoke are: \emph{Authority} and \emph{Subversion} with an undertone of \emph{Harm}, \emph{Betrayal}, and \emph{Care}. Both pairs of \emph{Authority} and \emph{Subversion}, and \emph{Care} and \emph{Harm}, are diametrically opposed MFs, which may explain why they believe that vaccination is necessary for others, but not themselves.

The {\underline{\sc Mandate Debaters}} VHP includes $86,306$ users (11\%) who discuss the civil rights issues surrounding mandating vaccination against COVID-19. 
These users heavily adopt the VHFs that that everyone should make their own informed decisions about COVID-19 vaccines and that people should not be chastised on whether they decide to avoid the vaccine, but otherwise debate whether vaccine mandates are ever appropriate. 
They overwhelmingly adopt the VHF that all healthcare workers should be vaccinated against COVID-19, and that refusing the COVID-19 vaccine puts the lives of others at risk, but also adopt erosion of trust framings which surround perceived issues with COVID-19 vaccines, such as their concern that the AstraZeneca vaccine may cause blood clots. The predominant
the MFs the VHFs they evoke are \emph{Harm} and \emph{Subversion}, with an additional focus on \emph{Fairness} and \emph{Authority}. 
They have the highest adoption of \emph{Fairness} of all the VHPs, which aligns with their focus on what is fair with regard to COVID-19 vaccine mandates. 

The {\underline{\sc Misinformers}} VHP includes $37,906$ users (5\%) who aggressively propagate COVID-19 vaccine misinformation. 
These users intensely adopt VHFs containing misinformation or eroding trust in vaccines, and completely reject VHFs about vaccine mandates.
They are entirely demotivated to take the COVID-19 vaccine because they believe that it is unnecessary, since the survival rate of COVID-19 is 99.99\%, and that the vaccine does not provide immunity. 
They also believe that the COVID-19 vaccine is an unsafe poison, that the vaccine is experimental and should not be used for children, and that the vaccine has not been sufficiently tested.
These users evoke VHFs with MFs that clearly align with those evoked by the {\sc Undecided}, and entirely adopted by the {\sc Demotivated}.
Espousing these MFs corresponds to the undue influence of the {\sc Misinformers}, leading to the propagation of COVID-19 vaccine misinformation to a wider audience, which we can see is often adopted by both the {\sc Undecided} and the {\sc Demotivated}. 

The {\underline{\sc Conspirators}} VHP includes $26,822$ users (3\%) who believe in COVID-19 vaccine conspiracy theories.
These users solely adopt VHFs referring to conspiracy theories, e.g. the vaccine contains a neurotoxin; or the mRNA vaccine is gene therapy which will change your DNA; or the government hides vaccine safety information, and that the vaccine itself contains the virus. 
The VHFs they adopt have MFs such as \emph{Degradation}, \emph{Harm}, \emph{Subversion}, and \emph{Authority}, and they belong to the only VHP which has heavy focus on \emph{Degradation}. 

The {\underline{\sc Concerned}} VHP includes $67,845$ users (8\%) who follow closely the science  regarding the COVID-19 vaccines, maintaining high vaccine literacy, but still have some minor concerns with the COVID-19 vaccines, manifested in their adoption of VHFs that erode trust in vaccines.
These users adopt VHFs which focus on the ability to reduce mortality by the COVID-19 vaccine, believing that the vaccines protect against grave forms of COVID-19 and are absolutely necessary for those at-risk.
They believe that the COVID-19 vaccines will protect against emerging variants, but are generally concerned with some specific VHFs, such as that the AstraZeneca vaccine may cause blood clots, the belief that the vaccines do not provide immunity, or the concern that the vaccine may not have been tested for long enough time yet.  
They adopt VHF with MFs such as \emph{Authority}, \emph{Care}, and \emph{Loyalty}.

The {\underline{\sc Promoters}} VHP includes $135,933$ users (17\%) who actively promote the role of the COVID-19 vaccines for public health. 
These users are characterized by overwhelming adoption and propagation of VHFs that build trust in vaccines, and the framings they evoke highlight their vaccine literacy.
These users overwhelmingly accept that vaccination is key in protecting yourself and others against COVID-19, that vaccination protects against severe COVID-19, that the vaccines will protect against emerging variants, and that the government has provided plenty of vaccine safety information. 
They also have a secondary focus on mythbusting through rejecting misinformation and trust-eroding framings, such as rejecting that the COVID-19 vaccine has not been sufficiently tested.
These users also adopt many VHFs that build trust in vaccines, while the VHFs they evoke have the MFs of \emph{Care}, \emph{Authority}, and \emph{Loyalty}.  

The {\underline{\sc Motivators}} VHP includes $131,717$ users (16\%) who specifically try to motivate users to get vaccinated. 
These users are similar to the {\sc Promoters},  widely adopting of VHFs for building trust in vaccines, e.g.  the COVID-19 vaccines protect against the emerging variants, or the vaccines trigger your body to naturally create immunity more reliably than getting COVID-19, and that the COVID-19 vaccines have been tested, tracked, and are safe. 
They also reject misinformation and VHFs eroding trust in vaccines.
These users have high vaccine literacy, and the VHFs they evoke have the MFs of \emph{Care} and \emph{Authority}.

The {\underline{\sc Motivated}} VHP includes $51,144$ users (6\%) who share stories of themselves and others getting vaccinated against COVID-19, including reassurance that the side effects are very minor and that the vaccines are extremely safe.
They adopt trust-building framings such as that the lingering effects and risks of COVID-19 are much worse than the minor side effects of getting vaccinated.
The VHF they evoke have the MFs of \emph{Care} and \emph{Authority}.



\section{Experimental Results}
\label{sec:results}

To evaluate the VHPs, we performed an external evaluation by sampling pairs of users and judging if these users should belong in the same or a different cluster based on the content of their tweets.
We selected the $50$ users for each VHP having user vector representations that are closest to
the VHP's  centroid vector.
For each of the $9$ VHPs, we then sampled from these $50$ users without replacement $20$ pairs of users from the same VHP and $20$ pairs of users from different VHPs, obtaining a total of $360$ pairs of users.
This approach ensured that bias was removed from the manual evaluation, such that each pair of users had a 50\% chance of being from the same VHP and a 50\% chance of being from a different VHP.
Language experts were tasked with judging if each pair was from the \emph{same} VHP cluster or from a \emph{different} VHP cluster by inspecting the content of the tweets of each pair of users.
We then compared these judgements to the VHP assignments obtained by k-means clustering. 
K-means clustering of $k=9$ produced 
a Rand index of $0.840$ 
and a Fowlkes-Mallows index of $0.847$ \cite{fowlkes-mallows}, with Precision of $0.815$ and Recall of $0.880$.
The random-clustering baseline had $0.5$ for the 
Rand index, the Fowlkes-Mallows index, Precision, and Recall.
The comparison with the random-clustering baseline demonstrates quantitatively that the $9$ VHP clusters are better-than-random. Moreover, our results show that 
the $9$ VHPs are of high quality, since users were $7.3$ times more likely to be identified as similar if they were from the same VHP and $4.0$ times more likely to be identified as different if they were from different VHPs.

\begin{table*}[th]
\centering
\small
\begin{tabular}{lrrrrrrrrr}
\toprule
System & Macro F1 & Macro P & Macro R & Accept F1 & Accept P & Accept R & Reject F1 & Reject P & Reject R \\
\midrule
{\sc StanceId-Baseline} & 69.1 & 68.8 & 69.5 & 81.0 & \textbf{79.3} & 82.8 & 57.2 & 58.2 & 56.2 \\
{\sc StanceId} & 72.4 & 69.6 & 75.4 & 80.6 & 77.1 & 84.5 & 64.1 & 62.1 & \textbf{66.3} \\
{\sc StanceId-Morality} & \textbf{75.2} & \textbf{73.0} & \textbf{77.9} & \textbf{83.6} & 77.8 & \textbf{90.5} & \textbf{66.8} & \textbf{68.3} & 65.4 \\
\bottomrule
\end{tabular}
\caption{Framing Stance Recognition results on the {\sc CoVaxFrames} test collection.}
\label{tb:results}
\end{table*}

The discovery of the VHPs was made possible by the recognition of the tweets that evoke the VHFs that we identified and the stance of the tweet authors. Therefore, we also evaluated the quality of stance recognition.
Stance recognition performance towards VHFs in the test collection of {\sc CoVaxFrames} was evaluated on three systems: 
(1) the {\sc StanceId-Baseline} system; 
(2) the {\sc StanceId} system; and 
(3) the {\sc StanceId-Morality} system.
The {\sc StanceId-Baseline} system utilizes the "[CLS]" embedding from COVID-Twitter-BERT-v2 as the framing stance recognition input embedding $z$. 
The {\sc StanceId} system utilizes Lexical, Emotion, and Semantic Graph Attention Networks to produce the framing stance recognition input embedding $z$ \cite{covid-misinfo-stance}. 
The {\sc StanceId-Morality} system, described in Section~\ref{sec:model} and illustrated in Figure~\ref{fig:architecture}, utilizes Lexical, Emotion, and Semantic Graph Attention Networks along with Hopfield Pooling of Moral Foundations to perform framing stance recognition.
Hyperparameters were selected based on initial experiments on the training and development collections of {\sc CoVaxFrames}. 
All system hyperparameters follow those of \citet{covid-misinfo-stance}, 
while {\sc StanceId-Morality} also performs Hopfield Pooling with Moral Foundations with $p=6$, has a GAT hidden size $F=32$, and $d=3$ stacked GAT layers.
All systems follow the same training schedule: $10$ epochs, a linearly decayed learning rate of $5e-4$ with a warm-up for $10\%$ of training steps, and an attention drop-out rate of $10\%$. 
Results are provided in Table~\ref{tb:results}.

Performance was determined based on Precision (P), Recall (R), and F$_1$\footnote{F$_1$ is defined as $F_1 =2 \times P \times R/(P+R)$} score for detecting the \emph{Accept} and \emph{Reject} values of stance. 
We also compute a Macro averaged Precision, Recall, and F$_1$ score. 
The {\sc StanceId-Baseline} system produced a Macro F$_1$ score of $69.1$, which demonstrates the advantage of pre-training BERT on domain-specific COVID-19 tweets and fine-tuning stance recognition systems. 
The {\sc StanceId} system produced a Macro F$_1$ score of $72.4$, which indicates that integrating Lexical, Emotional, and Semantic Graphs improves stance recognition. 
The {\sc StanceId-Morality} system produced a Macro F$_1$ score of $75.2$, supporting our hypothesis that MFs play a key role in detecting tweets which evoke VHFs along with recognizing \emph{acceptance} and \emph{rejection} of VHFs. 
The results also show that detecting \emph{rejection} of VHFs is more difficult than detecting \emph{acceptance}.

Improvements in stance recognition for the {\sc StanceId} system are driven by results for the \emph{Reject} stance. 
The \emph{Reject} stance has the fewest number of [tweet, VHF] pairs, with only 2,327 instances in a dataset of 14,180 [tweet, VHF] pairs. 
The {\sc StanceId} system overcomes this resource constraint by integrating additional Lexical, Emotion, and Semantic information. 
Stance recognition is further improved by the {\sc StanceId-Morality} system for both the \emph{Accept} and \emph{Reject} stance values.
The {\sc StanceId-Morality} system clearly benefits from integrating MF resources with the Hopfield pooling approach, which provides the best results on recognizing both \emph{acceptance} and \emph{rejection}.

\section{Ethics Statement}
\label{sec:ethics}
Accurate vaccine hesitancy profiling at scale has the potential to enable public health researchers to design customized interventions to target users most likely to be convinced to vaccinate. 
Public health outreach could become much more personalized, directly addressing the themes, concerns, and moral priorities held by users on Twitter.
Potential downsides to the approach outlined in this paper include the mistaken assignment of Twitter users to hesitancy profiles, due to sarcasm, jokes, or untruthful postings, which may be difficult for our system to recognize. 
Additionally, many users may change their stance and remove or rebut their own tweets over time. 
The downside of mistaken user hesitancy profiling is minimal, as we expect our system to be used by public health practitioners when developing their interventions. Questionnaires that precede the application
of an intervention would filter out Twitter for incorrect hesitancy profiles. 

Our data collection process was reviewed and approved by the Institutional Review Board at  
the University of Texas at Dallas.
All tweets collected were public, and only the tweet IDs and annotations will be shared, such that others must go through the approval process to use the data.
We expect {\sc CoVaxFrames} would become a valuable resource for identifying Vaccine Hesitancy Framings, and recognizing the \emph{stance} each Twitter user has towards those framings. 
While we believe that the annotation quality of {\sc CoVaxFrames} is high (0.67 Cohen's Kappa score), mistakes in judgements of \emph{stance} are likely due to difficult complex phenomena, such as sarcasm. 
We believe that such potential misjudgments are rare and thus minimally impact the quality of the hesitancy profiles.

\section{Conclusion}
\label{sec:concl}

In this paper we described a novel methodology for recognizing Vaccine Hesitancy Profiles (VHPs), applied to the COVID-19 vaccines. This methodology relies on the identification of how people frame their vaccine hesitancy, what Moral Foundations are implied by their Vaccine Hesitancy Framings (VHFs), and what stance the Tweet authors have towards  {\sc CoVaxFrames}. By considering the ontological commitments of the VHFs from {\sc CoVaxFrames} we derived nine VHPs of 805,336 Twitter users having a stance towards some {\sc CoVaxFrames}. The interpretation of the VHPs revealed that 22\% of these users are {\sc Undecided}; 11\% are  {\sc Demotivated}; 11\% are  {\sc Mandate Debaters}; 5\% are  {\sc Misinformers}; 3\% are  {\sc Conspirators}; 8\% are  {\sc Concerned}; 17\% are  {\sc Promoters}; 16\% are  {\sc Motivators}; and 6\% are  {\sc Motivated}.


%

\bibliography{aaai21}

\begin{thebibliography}{34}
\providecommand{\natexlab}[1]{#1}
\providecommand{\url}[1]{\texttt{#1}}
\providecommand{\urlprefix}{URL }
\expandafter\ifx\csname urlstyle\endcsname\relax
  \providecommand{\doi}[1]{doi:\discretionary{}{}{}#1}\else
  \providecommand{\doi}{doi:\discretionary{}{}{}\begingroup
  \urlstyle{rm}\Url}\fi

\bibitem[{Auxier and Anderson(2021)}]{pew1}
Auxier, B.; and Anderson, M. 2021.
\newblock Social media use in 2021.
\newblock \emph{Pew Research Center} .

\bibitem[{Beaulieu et~al.(1997)Beaulieu, Gatford, Huang, Robertson, Walker, and
  Williams}]{bm25}
Beaulieu, M.~M.; Gatford, M.; Huang, X.; Robertson, S.; Walker, S.; and
  Williams, P. 1997.
\newblock Okapi at TREC-5.
\newblock In \emph{The Fifth Text REtrieval Conference (TREC-5)}, 143--165.

\bibitem[{Cambria et~al.(2018)Cambria, Poria, Hazarika, and Kwok}]{senticnet-5}
Cambria, E.; Poria, S.; Hazarika, D.; and Kwok, K. 2018.
\newblock SenticNet 5: Discovering Conceptual Primitives for Sentiment Analysis
  by Means of Context Embeddings.
\newblock In \emph{AAAI Conference on Artificial Intelligence}.

\bibitem[{Chong and Druckman(2007)}]{Druckman}
Chong, D.; and Druckman, J.~N. 2007.
\newblock Framing Theory.
\newblock \emph{Annual Review of Political Science} 10: 103--126.

\bibitem[{Clevert, Unterthiner, and Hochreiter(2016)}]{elu}
Clevert, D.-A.; Unterthiner, T.; and Hochreiter, S. 2016.
\newblock Fast and Accurate Deep Network Learning by Exponential Linear Units
  (ELUs).

\bibitem[{Das et~al.(2007)Das, Datar, Garg, and Rajaram}]{lsh}
Das, A.~S.; Datar, M.; Garg, A.; and Rajaram, S. 2007.
\newblock Google News Personalization: Scalable Online Collaborative Filtering.
\newblock In \emph{Proceedings of the 16th International Conference on World
  Wide Web}, WWW '07, 271–280. New York, NY, USA: Association for Computing
  Machinery.

\bibitem[{Devlin et~al.(2019)Devlin, Chang, Lee, and Toutanova}]{bert}
Devlin, J.; Chang, M.-W.; Lee, K.; and Toutanova, K. 2019.
\newblock {BERT}: Pre-training of Deep Bidirectional Transformers for Language
  Understanding.
\newblock In \emph{Proceedings of the 2019 Conference of the North {A}merican
  Chapter of the Association for Computational Linguistics: Human Language
  Technologies, Volume 1 (Long and Short Papers)}, 4171--4186. Association for
  Computational Linguistics.

\bibitem[{Du et~al.(2017)Du, Xu, Song, and Tao}]{TaoCui}
Du, J.; Xu, J.; Song, H.-Y.; and Tao, C. 2017.
\newblock Leveraging machine learning-based approaches to assess human
  papillomavirus vaccination sentiment trends with Twitter data.
\newblock \emph{BMC Medical Informatics and Decision Making} 17.

\bibitem[{Entman(2004)}]{Entman}
Entman, R.~M. 2004.
\newblock \emph{Projections of Power: Framing News, Public Opinion, and U.S.
  Foreign Policy}.
\newblock University of Chicago Press.

\bibitem[{Field et~al.(2018)Field, Kliger, Wintner, Pan, Jurafsky, and
  Tsvetkov}]{field-etal-2018-framing}
Field, A.; Kliger, D.; Wintner, S.; Pan, J.; Jurafsky, D.; and Tsvetkov, Y.
  2018.
\newblock Framing and Agenda-setting in {R}ussian News: a Computational
  Analysis of Intricate Political Strategies.
\newblock In \emph{Proceedings of the 2018 Conference on Empirical Methods in
  Natural Language Processing}, 3570--3580. Association for Computational
  Linguistics.

\bibitem[{Foundation(1999)}]{lucene}
Foundation, A.~S. 1999.
\newblock Apache Lucene.
\newblock \url{https://github.com/apache/lucene}.

\bibitem[{Fowlkes and Mallows(1983)}]{fowlkes-mallows}
Fowlkes, E.~B.; and Mallows, C.~L. 1983.
\newblock A Method for Comparing Two Hierarchical Clusterings.
\newblock \emph{Journal of the American Statistical Association} 78(383):
  553--569.

\bibitem[{Garett and Young(2021)}]{Renee_Sean}
Garett, R.; and Young, S.~D. 2021.
\newblock Online misinformation and vaccine hesitancy.
\newblock \emph{Translational Behavior Medicine} 11: 2194--2199.

\bibitem[{Haidt and Graham(2007)}]{Haidt_Graham}
Haidt, J.; and Graham, J. 2007.
\newblock When morality opposes justice: Conservatives have moral intuitions
  that liberals may not recognize.
\newblock \emph{Social Justice Research} 20: 98--116.

\bibitem[{Haidt and Joseph(2004)}]{Haidt_Joseph}
Haidt, J.; and Joseph, C. 2004.
\newblock Intuituve ethics: How innately prepared intuitions generate
  culturally variable virtues.
\newblock \emph{Social Justice Research} 133: 55--66.

\bibitem[{Kata(2010)}]{Kata}
Kata, A. 2010.
\newblock A postmodern Pandora's box: Anti-vaccination misinformation on the
  Internet.
\newblock \emph{Vaccine} 28: 1709--16.

\bibitem[{Kingma and Ba(2015)}]{adam}
Kingma, D.~P.; and Ba, J. 2015.
\newblock Adam: {A} Method for Stochastic Optimization.
\newblock In \emph{3rd International Conference on Learning Representations,
  {ICLR} 2015, San Diego, CA, USA, May 7-9, 2015, Conference Track
  Proceedings}.

\bibitem[{Larson et~al.(2015)Larson, Jarrett, Schulz, Chadhuri, Zhou, Dube,
  Shuster, MacDonald, Wilson, and on~Vacine~Hesitancy}]{SAGE}
Larson, H.~J.; Jarrett, C.; Schulz, W.~S.; Chadhuri, M.; Zhou, Y.; Dube, E.;
  Shuster, M.; MacDonald, N.~E.; Wilson, R.; and on~Vacine~Hesitancy, S. W.~G.
  2015.
\newblock Measuring Vaccine Hesitancy, the Development of a Survey Tool.
\newblock \emph{Vaccine} 33: 4165--4175.

\bibitem[{Lloyd(1982)}]{kmeans}
Lloyd, S. 1982.
\newblock Least squares quantization in PCM.
\newblock \emph{IEEE Transactions on Information Theory} 28(2): 129--137.

\bibitem[{Luo, Zimet, and Shah(2019)}]{TenYears}
Luo, X.; Zimet, G.; and Shah, S. 2019.
\newblock A natural language processing framework to analyze the opinions on
  HPV vaccination reflected in twitter over 10 years (2008-2017).
\newblock \emph{Human Vaccines and Immunotherapeutics} 33: 1496--1504.

\bibitem[{Macdonald(2015)}]{3C}
Macdonald, N. 2015.
\newblock Vaccine hesitancy: Definition, scope and determinants.
\newblock \emph{Vaccine} 32.

\bibitem[{McHugh(2012)}]{kappa}
McHugh, M.~L. 2012.
\newblock Interrater reliability: the kappa statistic.
\newblock \emph{Biochemia medica} 22(3): 276--282.

\bibitem[{Müller, Salathé, and Kummervold(2020)}]{covid-twitter-bert}
Müller, M.; Salathé, M.; and Kummervold, P.~E. 2020.
\newblock COVID-Twitter-BERT: A Natural Language Processing Model to Analyse
  COVID-19 Content on Twitter.
\newblock \emph{https://arxiv.org/abs/2005.07503} .

\bibitem[{Nenkova and Passonneau(2004)}]{pyramid}
Nenkova, A.; and Passonneau, R. 2004.
\newblock Evaluating Content Selection in Summarization: The Pyramid Method.
\newblock In \emph{Proceedings of the Human Language Technology Conference of
  the North {A}merican Chapter of the Association for Computational
  Linguistics: {HLT}-{NAACL} 2004}, 145--152. Association for Computational
  Linguistics.

\bibitem[{Nogueira and Cho(2020)}]{bert-rerank}
Nogueira, R.; and Cho, K. 2020.
\newblock Passage Re-ranking with BERT.

\bibitem[{Ramsauer et~al.(2020)Ramsauer, Sch{\"{a}}fl, Lehner, Seidl, Widrich,
  Gruber, Holzleitner, Pavlovic, Sandve, Greiff, Kreil, Kopp, Klambauer,
  Brandstetter, and Hochreiter}]{HopfieldNets}
Ramsauer, H.; Sch{\"{a}}fl, B.; Lehner, J.; Seidl, P.; Widrich, M.; Gruber, L.;
  Holzleitner, M.; Pavlovic, M.; Sandve, G.~K.; Greiff, V.; Kreil, D.~P.; Kopp,
  M.; Klambauer, G.; Brandstetter, J.; and Hochreiter, S. 2020.
\newblock Hopfield Networks is All You Need.
\newblock \emph{CoRR} .

\bibitem[{Rao et~al.(2020)Rao, Morstatter, Hu, Chen, Burghardt, Ferrara, and
  Lerman}]{Lerman20}
Rao, A.; Morstatter, F.; Hu, M.; Chen, E.; Burghardt, K.; Ferrara, E.; and
  Lerman, K. 2020.
\newblock Political Partisanship and Anti-Science Attitudes in Online
  Discussions about COVID-19 (Preprint).
\newblock \emph{Journal of Medical Internet Research} 23.

\bibitem[{Rossen et~al.(2019)Rossen, Hurlstone~Angeli, Dunlop, and
  Lawrence}]{Rossen}
Rossen, I.; Hurlstone~Angeli, M.~J.; Dunlop, P.~D.; and Lawrence, C. 2019.
\newblock Accepters, fence sitters, or rejecters: Moral profiles of vaccination
  attitudes.
\newblock \emph{Social Science Medicine} 23--27.

\bibitem[{Roy and Goldwasser(2020)}]{roy-goldwasser-2020-weakly}
Roy, S.; and Goldwasser, D. 2020.
\newblock Weakly Supervised Learning of Nuanced Frames for Analyzing
  Polarization in News Media.
\newblock In \emph{Proceedings of the 2020 Conference on Empirical Methods in
  Natural Language Processing (EMNLP)}, 7698--7716. Online: Association for
  Computational Linguistics.

\bibitem[{Thorndike(1953)}]{elbow}
Thorndike, R.~L. 1953.
\newblock Who belongs in the family?
\newblock \emph{Psychometrika} 18(4): 267--276.
\newblock ISSN 1860-0980.

\bibitem[{Veličković et~al.(2018)Veličković, Cucurull, Casanova, Romero,
  Liò, and Bengio}]{gat}
Veličković, P.; Cucurull, G.; Casanova, A.; Romero, A.; Liò, P.; and Bengio,
  Y. 2018.
\newblock Graph Attention Networks.
\newblock In \emph{International Conference on Learning Representations}.

\bibitem[{Weinzierl and
  Harabagiu(2020{\natexlab{a}})}]{covid-hopfield-pool-event}
Weinzierl, M.; and Harabagiu, S. 2020{\natexlab{a}}.
\newblock {HLTRI} at {W}-{NUT} 2020 Shared Task-3: {COVID}-19 Event Extraction
  from {T}witter Using Multi-Task Hopfield Pooling.
\newblock In \emph{Proceedings of the Sixth Workshop on Noisy User-generated
  Text (W-NUT 2020)}, 530--538. Association for Computational Linguistics.

\bibitem[{Weinzierl and Harabagiu(2020{\natexlab{b}})}]{ser4eqnova}
Weinzierl, M.; and Harabagiu, S.~M. 2020{\natexlab{b}}.
\newblock The University of Texas at Dallas HLTRI’s Participation in EPIC-QA:
  Searching for Entailed Questions Revealing Novel Answer Nuggets.
\newblock In \emph{Thirteenth Text Analysis Conference}, volume~13. Text
  Analysis Conference.

\bibitem[{Weinzierl, Hopfer, and Harabagiu(2021)}]{covid-misinfo-stance}
Weinzierl, M.; Hopfer, S.; and Harabagiu, S. 2021.
\newblock Misinformation Adoption or Rejection in the Era of COVID-19.
\newblock In \emph{Proceedings of the International AAAI Conference on Web and
  Social Media (ICWSM)}. {AAAI} Press.

\end{thebibliography}

\end{document}